\documentclass[letterpaper,10pt,conference]{ieeeconf}
\IEEEoverridecommandlockouts
\usepackage{cite}
\usepackage{amsmath,amssymb}
\usepackage{graphicx}
\usepackage{booktabs}
\usepackage{textcomp}
\usepackage{xcolor}
\usepackage[hidelinks]{hyperref}

\hypersetup{
  pdftitle={TACIT: Tactile Contact Supervision for Spatial Attention in Dexterous Manipulation},
  pdfauthor={Yanhou Lai, Fucai Zhu, Ruiqiang Wang, Koichi Hashimoto}
}

\def\BibTeX{{\rm B\kern-.05em{\sc i\kern-.025em b}\kern-.08em T\kern-.1667em\lower.7ex\hbox{E}\kern-.125emX}}

\begin{document}

\title{\LARGE \bf
TACIT: Tactile Contact Supervision for\\
Spatial Attention in Dexterous Manipulation
}

\author{Yanhou Lai, Fucai Zhu, Ruiqiang Wang, and Koichi Hashimoto%
\thanks{All authors are with the Graduate School of Information Sciences,
Tohoku University, Sendai 980-8579, Japan.}%
\thanks{This work has been submitted to the IEEE for possible publication.
Copyright may be transferred without notice, after which this version may no
longer be accessible.}}

\maketitle
\thispagestyle{empty}
\pagestyle{empty}

\begin{abstract}

Visuomotor policies trained from a few demonstrations may reproduce demonstrated trajectories without reliably following changes in object position. Existing approaches with explicit attention typically obtain spatial priors from human annotation or visual models. We introduce TACIT (tactile contact informs attention), which uses measured tactile contacts from teleoperated demonstrations to supervise spatial attention without additional point annotation. Gaussian targets over preceding camera point clouds supervise an attention head whose pooled output conditions a visuotactile diffusion policy. Targets are used only during training; tactile observations remain inputs at inference. In the primary real-robot benchmark, with ten demonstrations per task and five demonstrated placement regions, TACIT achieves 66.7 \% success on ball placement and 73.3 \% on peg insertion, compared with 10.0 \% and 20.0 \% for input-matched 3D visuotactile fusion and 20.0 \% and 43.3 \% for vision-only DP3. TACIT enters the 150\,mm palm-to-object approach region within 12 seconds in all 30 trials per task; all remaining failures occur after arrival. Across three training seeds on real ball and simulated peg, TACIT outperforms input-matched fusion and an architecture-matched control without explicit attention supervision, supporting the contribution of supervision beyond branch capacity. Pre-contact and contact-time supervision show no consistent ordering. These results demonstrate that measured tactile contact provides effective spatial supervision for approach behavior from few demonstrations within the evaluated workspace.

\end{abstract}
\section{Introduction}\label{sec:introduction}

Learning from few demonstrations reduces collection effort but limits spatial variation, allowing a policy to reproduce demonstrated motion without grounding the object in the current scene. We observe this directly: in 87 of 92 failed baseline trials on two real-world tasks, the hand ends at the wrong location rather than failing to grasp. Trajectory analysis also shows that the tactile-fusion baseline weakly follows object displacement, whereas TACIT tracks it closely (Sec.~\ref{sec:real-results}).

\begin{figure}[t]
\centering
\includegraphics[width=\columnwidth]{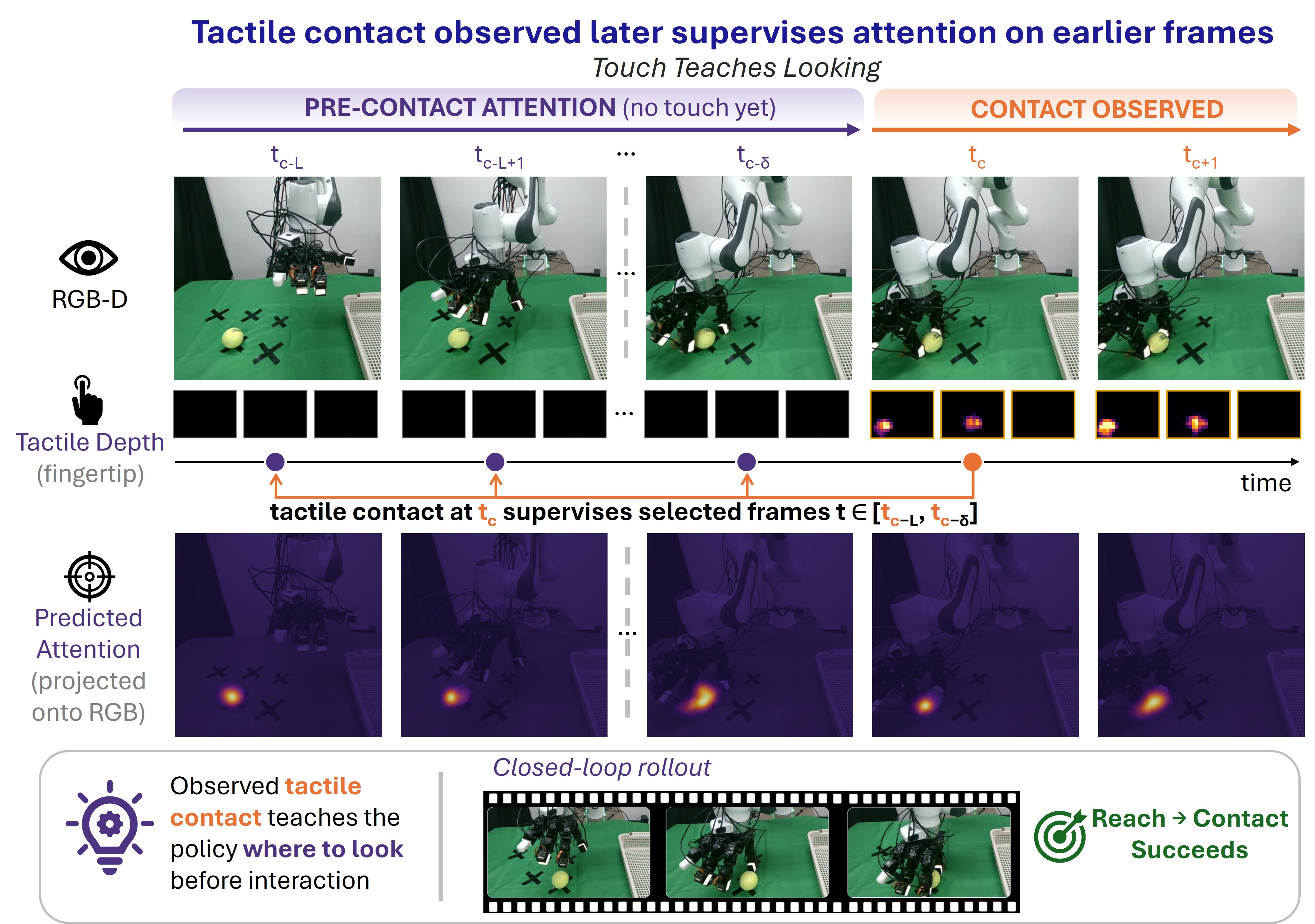}
\caption{TACIT overview. In the illustrated ball demonstration, contact at $t_c$ provides training-only supervision for spatial attention at earlier frames $t<t_c$. Attention is predicted on sampled camera points and projected onto RGB for visualization.}
\label{fig:tacit_policy_overview_handmade}
\end{figure}

Explicit spatial priors focus a policy on a small set of points. Prior work obtains such priors from human annotation \cite{p3po,guidedattention}, vision foundation models \cite{skil,gendp}, or task-driven keypoint selection \cite{atk}.

As illustrated in Fig.~\ref{fig:tacit_policy_overview_handmade}, tactile contact provides a physical spatial target without additional per-demonstration annotation, but it is observed only after the approach. TACIT retrospectively assigns the contact location observed later in the demonstration to preceding camera observations as training-only attention supervision.

Three properties characterize how this supervision is obtained and used:

\begin{itemize}
\item \textbf{Automatic label generation.} The labels require no manual point selection, object model, or pretrained semantic model for policy training or control. The tactile hardware is a fixed system cost rather than a per-demonstration annotation cost.
\item \textbf{Attention supervision on earlier frames.} Each target comes from a selected later contact, so the camera-only head learns to predict its location before that contact occurs. At inference, the head predicts attention from current camera points; tactile observations remain inputs to the policy encoder.
\item \textbf{Measured physical interaction targets.} TACIT reconstructs a physical interaction location from measured tactile contact and fingertip kinematics recorded during demonstration, without externally supplied object pose or spatial annotation.
\end{itemize}

With ten demonstrations per task, TACIT eliminates observed pre-arrival failures under the 150\,mm criterion in both primary benchmarks; all remaining failures occur after arrival. Three-seed comparisons on real ball and simulated peg confirm gains over input-matched fusion and architecture-matched controls.

\section{Related Work}\label{sec:related-work}

\textbf{Spatial priors for manipulation policies.} VIOLA \cite{viola} uses object proposals; P3-PO \cite{p3po} uses annotated points on one frame; SKIL \cite{skil} and GenDP \cite{gendp} derive semantic keypoints and fields from vision foundation models; ATK \cite{atk} selects task-driven keypoints from expert data; and GuidedAttention \cite{guidedattention} uses human clicks propagated by a tracker. Their guidance comes from visual models, human-specified points, or both.

\textbf{Demonstrator-derived attention supervision.} Human gaze recorded during demonstrations has trained visual-attention models for visuomotor imitation \cite{agil} and robot manipulation \cite{kimgaze}, and has guided policy features via an auxiliary loss \cite{gazeil}. TACIT instead derives a 3D policy-frame target from measured contact at the interaction location, without eye tracking.

\textbf{Point-cloud and 3D visuotactile policies.} Diffusion Policy \cite{diffusionpolicy} introduced conditional action denoising; our DP3 \cite{dp3} backbone conditions it on a compact point cloud. RISE \cite{rise} uses sparse 3D tokens, while ISS Policy \cite{isspolicy} adds training-only future-scene supervision without a tactile spatial target. 3D-ViTac \cite{3dvitac} fuses kinematically localized taxels and visual points, the representation family used by our input-matched baseline. TransDex \cite{transdex} combines related fusion with point-cloud pretraining.

\textbf{Contact-aware multimodal policies.} Visuo-Tactile Transformers \cite{vtt} form visual heatmaps using tactile feedback and cross-modal attention. TacImag \cite{tacimag} predicts tactile representations from vision and proprioception for tactile-free deployment. Contact-Grounded Policy \cite{cgp}, Reactive Diffusion Policy \cite{rdp}, AdapTac \cite{adaptac}, CAAT \cite{caat}, and Dream-Tac \cite{dreamtac} use tactile state or predicted tactile/force signals for control or fusion. FoAR \cite{foar} combines vision with force/torque sensing; RGB-S \cite{rgbs} projects kinematically localized tactile contacts into images as force-modulated saliency maps. TACIT instead uses later contacts to supervise camera-point attention on preceding observations during training.

\textbf{Grasp synthesis and contact maps.} UniDexGrasp \cite{unidexgrasp} uses ContactNet to guide grasp generation; DexDiffuser \cite{dexdiffuser} and DexGrasp-Diffusion \cite{dexgraspdiffusion} generate and evaluate dexterous grasps. TACIT instead uses measured demonstration contacts to supervise a closed-loop visuomotor policy.

\section{Method}\label{sec:method}

\begin{figure*}[t]
\centering
\includegraphics[width=\textwidth]{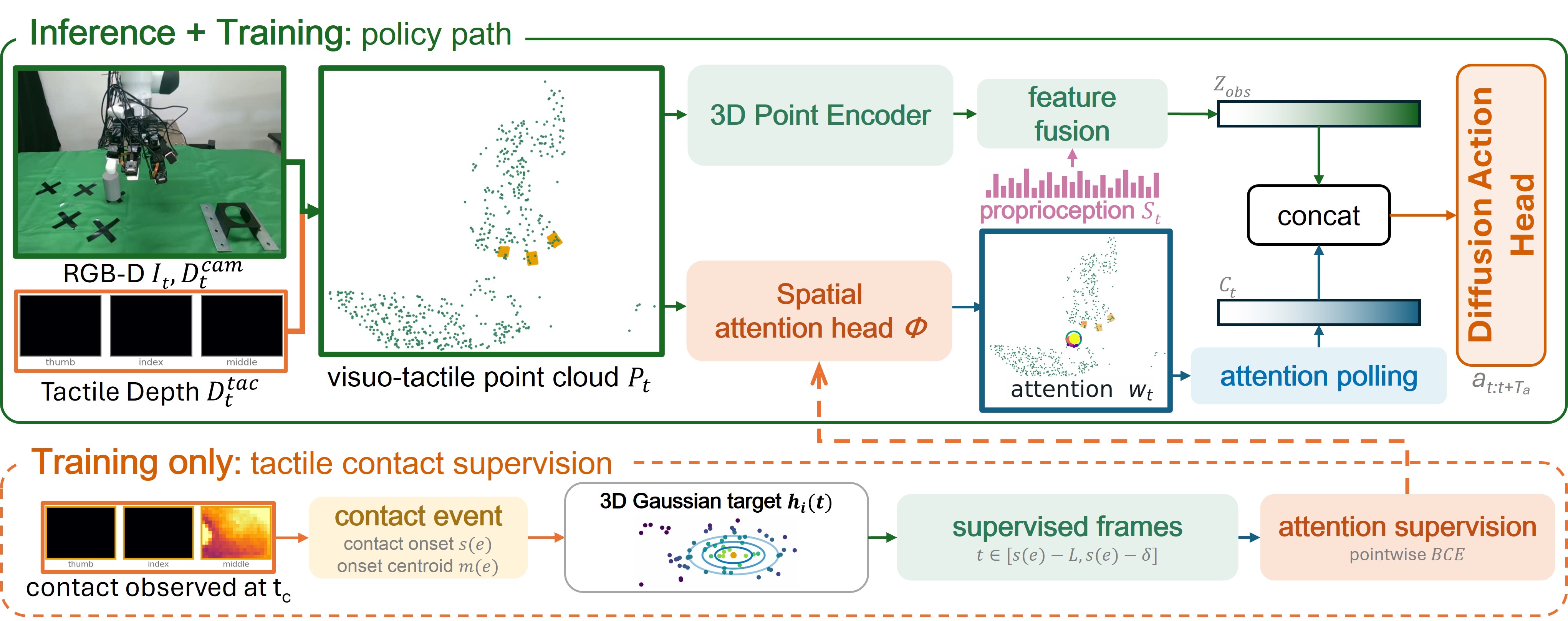}
\caption{TACIT architecture. An attention head over camera points supplies a pooled spatial feature alongside the shared visuotactile observation encoding. During training, Gaussian targets derived from tactile contact supervise the camera-point scores. Bimanual control instantiates one attention head per hand.}
\label{fig:tacit_policy_architecture}
\end{figure*}

TACIT has two components. First, recorded tactile contacts provide spatial targets for attention on earlier camera observations. Second, an attention head pools the current camera cloud into a feature that conditions the diffusion policy alongside the shared visuotactile encoding (Fig.~\ref{fig:tacit_policy_architecture}).

\subsection{Tactile Contact Supervision}\label{sec:contact-label}

\textbf{Contact event extraction.} At each frame, we select the finger with the largest number of active tactile cells and compute the centroid of its active taxels. Invalid contact frames contribute zero to the selected area signal, which is median-filtered to give $\tilde a_t$. Event detection combines contact validity with the following threshold conditions, using $\theta_{\mathrm{on}}$ 5\,cells on the robot and 20 in simulation to reflect their different contact-patch scales:
\begin{equation}
\begin{aligned}
\text{on:}\quad &\tilde a_t \ge \theta_{\mathrm{on}}\\
\text{off:}\quad &\tilde a_t < \theta_{\mathrm{off}},\quad \theta_{\mathrm{off}} = \theta_{\mathrm{on}}/2
\end{aligned}
\end{equation}
Opening requires valid contact; invalid frames also satisfy the off condition. Consecutive-frame requirements debounce transitions, and short event runs are filtered. Runs separated by at most 8\,frames are merged.

From the first eight valid frames of each event, we retain frames at or above the 80th percentile of active-cell count. Their taxel centroids are averaged using the active-cell counts as weights to obtain the onset centroid $m(e)$.

Contact centroids and camera points are both expressed in the robot base/world frame without an object pose. The calibrated $20 \times 15$ sensor grid is transformed by the fixed sensor-to-fingertip calibration and recorded fingertip pose; the RGB-D image is unprojected with the camera intrinsics and transformed by the calibrated camera extrinsic.

\textbf{Selecting the supervising contact event.} For an earlier frame $t$, we first select the contact event with the earliest onset $s(e)$ within the next $L$ frames and define its temporal lead:
\begin{equation}
\begin{aligned}
e(t) &= \operatorname*{arg\,min}_{e:\,1\leq s(e)-t\leq L}\,[s(e)-t]\\
\Delta(t) &= s(e(t))-t
\end{aligned}
\end{equation}
If no event falls within this horizon, we set $v_t=0$ and construct no target.

\textbf{Selecting supervised frames.} Let $\mathcal{Q}_t$ be the set of camera-point XYZ coordinates used for label generation at frame $t$. We retain a frame when the selected event is at least $\delta$ steps ahead and this cloud contains sufficient Gaussian target support. We use a Gaussian with spatial standard deviation $\sigma$. Its maximum value on the label-generation cloud defines the support score $\kappa_t$; the validity mask requires this score to reach $\eta$ (brackets denote indicators):
\begin{equation}
\begin{aligned}
\kappa_t &= \max_{q\in\mathcal{Q}_t}
  \exp\!\left(-\frac{\lVert q-m(e(t))\rVert^2}{2\sigma^2}\right),\\
v_t &= [\Delta(t)\geq\delta]\,[\kappa_t\geq\eta].
\end{aligned}
\end{equation}
We use a 25\,mm Gaussian width and a support threshold of 0.1. The selected frames lie 5--120 steps before the supervising contact onset, excluding the immediate-contact neighborhood. These fixed offsets correspond to 0.25--6.0\,s in hardware and 0.17--4.0\,s in simulation.

\textbf{Aligning targets with policy inputs.} For each retained frame, we evaluate the contact-centered Gaussian at the camera points sampled for policy training. For six-channel point $p_i(t)$ with XYZ coordinate $x_i(t)$, the target is
\begin{equation}
h_i(t) = \exp\!\left(
-\frac{\lVert x_i(t)-m(e(t))\rVert^2}{2\sigma^2}\right)
\end{equation}
Targets are thus aligned with the policy input while retaining the original frame selection; points nearer the contact centroid receive larger values.

\textbf{Auxiliary attention objective.} Let $\mathcal{T}=\{t:v_t=1\}$ denote the retained supervised frames and $\mathcal{V}_t$ the policy's sampled camera-point indices at timestep $t$. The pointwise attention loss over these frames is
\begin{equation}
\mathcal{L}_{\mathrm{attn}} =
\operatorname*{mean}_{t \in \mathcal{T},\, i \in \mathcal{V}_t}
\mathrm{BCE}\!\left(\operatorname{sigmoid}(\ell_i(t)),\, h_i(t)\right)
\end{equation}
This loss trains each camera point's score toward the contact-derived Gaussian. Frame validity masks only this loss; diffusion training retains every sampled demonstration frame. The joint objective is
\begin{equation}
\mathcal{L} = \mathcal{L}_{\mathrm{diff}}
+ \lambda_{\mathrm{att}}\,\mathcal{L}_{\mathrm{attn}},
\qquad \lambda_{\mathrm{att}} = 0.05
\end{equation}
\subsection{TACIT Policy}\label{sec:attention-policy}

\textbf{Spatial attention scores.} An MLP scores every point of the cloud with shared weights (6$\rightarrow$64$\rightarrow$64$\rightarrow$1). Masking restricts attention to camera points; tactile points still enter the shared observation encoder but never receive attention weight. For full six-channel point $p_i(t)$, the masked logit is
\begin{equation}
\ell_i(t) =
\begin{cases}
\phi(p_i(t)), & i \in \mathcal{V}_t,\\
-\infty, & i \notin \mathcal{V}_t,
\end{cases}
\end{equation}
During training, sigmoid-BCE treats these logits as independent pointwise contact-relevance scores. For policy conditioning, we instead normalize the same logits into spatial-pooling weights:
\begin{equation}
w_i(t) = \frac{\exp(\ell_i(t)/\tau)}
  {\sum_{j\in\mathcal{V}_t}\exp(\ell_j(t)/\tau)}, \qquad \tau=1
\end{equation}
The weights pool the six-channel camera points into a single feature:
\begin{equation}
\bar p_t = \sum_{i\in\mathcal{V}_t} w_i(t) p_i(t)
\end{equation}
The pooled feature gives the attended location and point attributes but does not retain absolute pointwise relevance or the spread of the weights. We therefore also compute the peak sigmoid score and an entropy term scaled by the full sampled-cloud size $N$:
\begin{equation}
\begin{aligned}
\rho_t &= \max_{i\in\mathcal{V}_t}\operatorname{sigmoid}(\ell_i(t))\\
H_t &= -\frac{\sum_{i\in\mathcal{V}_t}w_i(t)\log w_i(t)}{\log N}
\end{aligned}
\end{equation}
The denominator includes visual and tactile points, so entropy is scaled by the full cloud size rather than the number of camera points.

\textbf{Attention-conditioned diffusion policy.} The resulting 8-D summary (six-channel pooled feature, peak relevance and scaled entropy) is projected by $g$ to a 16-D conditioning feature and concatenated with the standard observation encoding:
\begin{equation}
c_t = g([\bar p_t,\rho_t,H_t]), \qquad
z_t = [z^{\mathrm{obs}}_t,c_t]
\end{equation}
For a single arm, the 87-D observation feature contains the 64-D point-cloud encoding and 23-D proprioception, yielding a 103-D diffusion-policy input after concatenation.

\textbf{Bimanual extension.} For bimanual handover the same construction is instantiated once for the giver and once for the receiver. Both heads score the shared camera cloud, each is supervised only on frames preceding that hand's next contact, and their two features are concatenated with the shared observation encoding.

\section{Experimental Setup}\label{sec:experimental-setup}

Our evaluation asks four questions. \textbf{Q1:} Does TACIT improve real-world task success, and is the gain concentrated in the approach? \textbf{Q2:} Does the hand follow the current object rather than a fixed workspace destination, and does the head localize the object before touch? \textbf{Q3:} Does TACIT retain an advantage across simulated tasks and at held-out positions within the demonstrated spatial support? \textbf{Q4:} Does the gain require explicit attention supervision, and does pre-contact timing add anything over contact-time supervision?

\subsection{Real-Robot Tasks and Hardware}\label{sec:real-setup}

The hardware comprises a Franka FR3 arm, LEAP hand \cite{leaphand}, three 9DTact \cite{9dtact} compliant fingertips (thumb, index, middle), and one front RealSense D455; the ring fingertip is unsensed. Demonstrations are collected by teleoperation with a learned hand-retargeting model using the ViHaTeleop pipeline \cite{vihateleop}. We collect ten demonstrations for each of two tasks.

\emph{Ball placement} requires grasping a ball from the table and releasing it into a basket. Releasing makes no new contact; the detector finds 1.2 events per demonstration, all in the grasp phase, so its supervision concerns approach and grasp. \emph{Peg insertion} requires grasping a peg and inserting it into a 3D-printed fixture. Its 2.7 detected contact events per demonstration test the same event-based labeling in a multi-contact sequence.

\subsection{Simulation Tasks}\label{sec:sim-setup}

\begin{figure*}[t]
\centering
\includegraphics[width=0.92\textwidth]{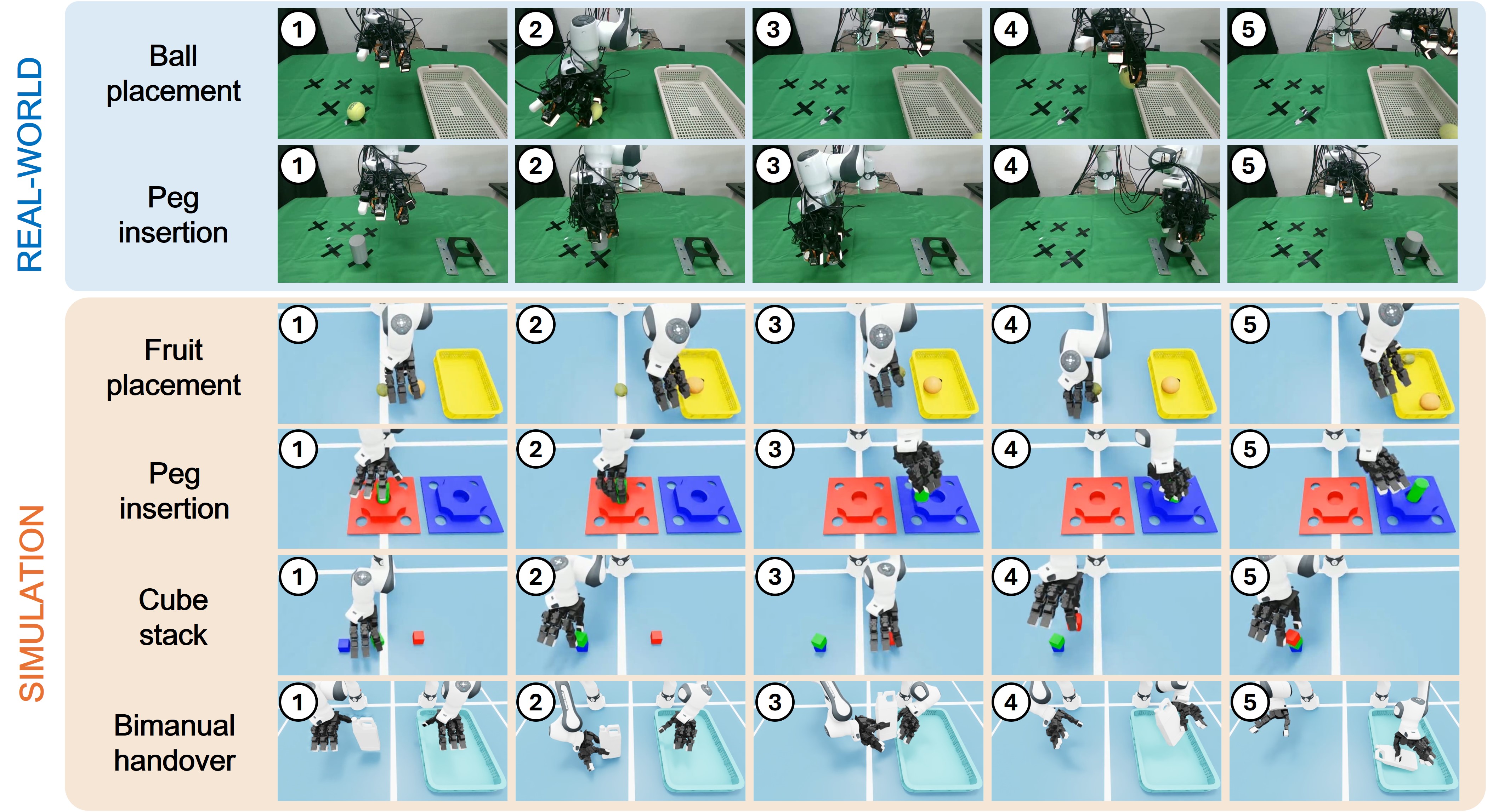}
\caption{Evaluation tasks. Real-robot ball placement and peg insertion provide the primary physical-policy results; simulation evaluates three single-arm tasks and bimanual handover. The real and simulated peg tasks are distinct instances. Numbers mark rollout stages.}
\label{fig:task_overview}
\end{figure*}

We evaluate fruit placement, peg insertion and cube stacking in IsaacLab as matched single-arm ablations, with bimanual handover reported separately as an architectural extension. Fig.~\ref{fig:task_overview} summarizes the real and simulated evaluation tasks. Single-arm cameras use the real D455's measured extrinsics and field of view; handover retains its two-arm camera configuration. Scripted experts provide two demonstrations per configuration: 50 for fruit and cube, and 18 for peg. Fruit and cube each combine five positions per movable object into 25 joint configurations ($10 \times 4$\,cm and $8 \times 6$\,cm support per object). Peg uses a paired $3 \times 3$ peg/hole grid over $12 \times 16$\,cm; handover uses five jug and two basket positions. Every policy is trained to epoch 3000 with the same label construction as on the robot; active-cell count selects simulated contact evidence. No policy is transferred between domains.

\subsection{Policy Baselines and Training}\label{sec:policy-setup}

Three policies differ in observation and supervision; Table~\ref{tab:1} summarizes their single-arm inputs.

\begin{table}[t]
\centering
\caption{Single-arm policy inputs and attention supervision.}
\label{tab:1}
\footnotesize
\setlength{\tabcolsep}{3pt}
\begin{tabular*}{\columnwidth}{@{}l@{\extracolsep{\fill}}ccc@{}}
\toprule
 & Point cloud & Tactile & Attention \\
\midrule
A1 DP3 & 1500 camera & no & no \\
A2 3D VT fusion & 600 camera + 900 tactile & yes & no \\
A3 TACIT (ours) & 600 camera + 900 tactile & yes & yes \\
\bottomrule
\end{tabular*}
\end{table}

A1 and A2 represent established vision-only and visuotactile policy baselines, respectively. A1 is a 3D diffusion policy \cite{dp3}: a PointNet \cite{pointnet} encoder over the point cloud, concatenated with proprioception, conditioning a 1-D conditional UNet with DDIM \cite{ddim} sampling---the standard point-cloud policy in this literature and TACIT's backbone. A2 is platform-matched shared-space 3D visuotactile fusion \cite{3dvitac}, using ViHaTeleop's pipeline, rather than a reproduction of the complete 3D-ViTac system. TACIT differs from A2 only by the attention head and its loss. A1 is a vision-only reference with a different camera-point budget, so A1--A2 does not isolate tactile fusion. A2 and TACIT share their camera and tactile point budgets, isolating the addition of contact-supervised attention. Handover uses 2,400 camera points for A1 and 600 camera plus 1,800 tactile points for A2 and TACIT.

A2 and TACIT use non-color six-channel points comprising XYZ, force, and binary visual/tactile modality masks in the robot base frame. The attention MLP scores each raw point independently. Since force and the two modality-mask channels are constant on visual points, its visual score is primarily conditioned on base-frame XYZ rather than appearance or scene-context features. Pooling over the current cloud lets the fixed scoring function move the attended location with observed geometry. State and action are 23-D for one arm and 46-D for handover. Adam uses batch 16, learning rate $10^{-4}$, and 3000 epochs of 500 minibatches. One observation conditions 20 predicted steps; ten are executed before replanning. DDIM uses 100 training and ten inference steps \cite{ddim}. Hardware control runs at 20\,Hz, and IsaacLab runs at 30\,Hz.

\textbf{Ablation variants.} We introduce two architecture-matched controls on simulated peg insertion and real ball placement. A2.5 retains TACIT's attention branch and policy conditioning but sets $\lambda_{\mathrm{att}}=0$, removing only the explicit attention-supervision loss; the head is still trained through the action objective. TACIT-contact uses the same contact-derived Gaussian target as TACIT but applies it during detected contact intervals rather than to preceding observations. Contact-time frames are subsampled only when needed to keep the supervised-frame count within 10 \% of TACIT's, which controls total supervision count but not per-event allocation. A2 and all three attention variants are trained with seeds 1, 2, and 3 using the same fixed training subsets and hyperparameters. The primary ball and cross-task peg entries retain seed 1; the three-seed comparison includes these checkpoints without reevaluating them.

\textbf{Label coverage.} For TACIT's pre-contact supervision, the real datasets contain 12 ball and 27 peg contact events. Frame selection retains 1,187/3,925 ball frames (30.2 \%) and 1,623/4,591 peg frames (35.4 \%), covering all ten demonstrations per task. After alignment with policy inputs, target peaks fall below 0.1 in one retained ball frame and 60 retained peg frames; because validity is set before alignment, these frames remain supervised.

\subsection{Evaluation Protocol and Metrics}\label{sec:evaluation-protocol}

\textbf{Real robot.} Trials are scored from front-camera video: the ball must be released inside the basket, and the peg must remain inserted in the printed fixture. Each trial uses a 30-s policy-execution horizon, with success recorded if the task criterion is satisfied within that window. Insertion timeout is assigned only after successful arrival and grasp; pre-arrival failures therefore do not enter this downstream category. Each policy is evaluated in thirty trials per task: five object positions, two trials each, across three sessions under the final fixed controller and success criteria. The object is re-placed by hand between trials, so positions are seen but not reproduced. Primary evaluation positions lie 3.6--13.8\,mm from the nearest demonstrated position. The reported real-robot benchmark uses no held-out spatial split. The primary benchmarks use one epoch-3000 checkpoint per policy and task. Attention localization covers real ball (three seeds) and real peg (seed 1).

We separate the approach from completion with three diagnostics. A trial is within reach if the palm enters the 150\,mm grasp-approach region around the object during the first 12 seconds. A threshold-free complement is the minimum fingertip-center distance to the object over the same window and three fingers; its absolute value is task-dependent and is used only to compare policies. For spatial tracking, let $o_j$ denote the per-placement median object position assigned to trial $j$ and $q_j$ the palm at closest approach over the full rollout; for each horizontal axis we fit
\begin{equation}
q_j = \beta\, o_j + \alpha + \varepsilon_j
\end{equation}
and bootstrap 95\% intervals over the five placements. We use the slope $\beta$ as the grounding measure: $\beta=1$ means the hand follows object displacement, whereas $\beta=0$ means it approaches a fixed workspace position. The intercept $\alpha$ captures only a constant spatial offset between hand and object. Attention localization is measured by the distance from the geometrically located object to both the attention centroid and highest-weight camera point. We retain stored frames in the first 9\,s but stop each trial before its first tactile activation, defined as any taxel depth below 8.5\,mm. This evaluator is independent of the head and is never used for policy training or control.

\textbf{Simulation.} Success requires endpoint geometry to hold for ten consecutive 30\,Hz steps while objects are quasi-static: containment for fruit, relative pose for peg and cube, and confirmed transfer followed by basket containment for handover. The fixed-support benchmark contains 225 fruit, 81 peg, 225 cube and 90 handover rollouts per policy. For each checkpoint, evaluation seeds 0, 1 and 42 traverse the complete placement set three times while diffusion noise is redrawn at every inference call. We pool sweeps within evaluation seed. The cross-task benchmark reports mean and standard deviation across evaluation-seed rates; the peg mechanism controls instead report three independently trained checkpoints, each evaluated over the same 81 rollouts. Placements, not repeated stochastic rollouts, are the independent unit.

The held-out diagnostic evaluates positions absent from demonstrations but inside each task's training convex hull. Fruit and cube use 16 joint placement cases, with each active object inside its corresponding 2D placement hull and 28.3\,mm (fruit) or 31.6\,mm (cube) in XY from the nearest demonstrated placement. Peg and hole translate together over four held-out pairs separated by 50.0\,mm from the nearest demonstrated pair. Each distance is near the largest feasible nearest-demonstration separation within that task's placement hull. Three evaluation seeds with three stochastic sweeps per case produce 144 fruit/cube and 36 peg trials per policy. Fixed stage metrics include fruit lifts, peg displacement, and sequential cube progress. Fruit lifts are counted independently in either order. A cube passes a lift/stack stage if lifted by over 8 cm or stacked; the second-cube stage requires the first stack. Successful rollouts pass all cube stages. This tests whether the effect is confined to exact demonstrated coordinates, not workspace extrapolation.

\section{Results}\label{sec:results}

\subsection{Real-World Results}\label{sec:real-results}

\textbf{Task success and reaching (Q1).} TACIT leads both baselines in task success and reaches every trial in the primary benchmark, whereas baseline reach and success nearly coincide (Table~\ref{tab:2}).

Against input-matched A2, TACIT has higher observed task success at all five ball placements and four of five peg placements, indicating that the pooled gain is not driven by a single placement.

\begin{table}[t]
\centering
\caption{Real-world reaching and task success (30 trials per policy per task).}
\label{tab:2}
\footnotesize
\setlength{\tabcolsep}{3pt}
\begin{tabular}{lrrrr}
\toprule
 & Ball reach & Ball success & Peg reach & Peg success \\
\midrule
A1 DP3 & 8/30 & 6/30 & 14/30 & 13/30 \\
A2 3D VT fusion & 4/30 & 3/30 & 6/30 & 6/30 \\
\textbf{A3 TACIT (ours)} & \textbf{30/30} & \textbf{20/30} & \textbf{30/30} & \textbf{22/30} \\
\bottomrule
\end{tabular}
\end{table}

Manual review of the recorded front-camera trial videos classifies 87 of 92 baseline failures in the included sessions as wrong-location failures, compared with 0 of 18 TACIT failures. The geometric threshold independently classifies the same aggregate 87 of 92 baseline failures as pre-arrival, although the two criteria disagree on two individual ball trials.

Manual review further separates TACIT's 18 post-arrival failures: all ten ball failures reach the grasp region but fail to secure the ball, while the eight peg failures comprise four failed insertion attempts and four trials in which prolonged grasp-stabilization behavior after tactile contact delays insertion beyond the 30-s evaluation window.

On ball, median [interquartile range] fingertip-center distances for A1/A2/TACIT are 142.8 [68.3, 205.9], 128.3 [107.3, 200.0], and \textbf{51.2 [46.8, 54.6] mm}. On peg, they are 82.6 [44.5, 157.7], 134.9 [90.3, 183.4], and \textbf{50.2 [43.6, 55.0] mm}. The continuous distance measure shows the same separation without an additional threshold. Together, these results place the main gain in approach: most baseline failures precede arrival, while all TACIT failures follow it in the primary benchmark.

A2's lower success than A1 does not isolate tactile input because modality and visual sampling budget both change.

\begin{figure}[t]
\centering
\includegraphics[width=\columnwidth]{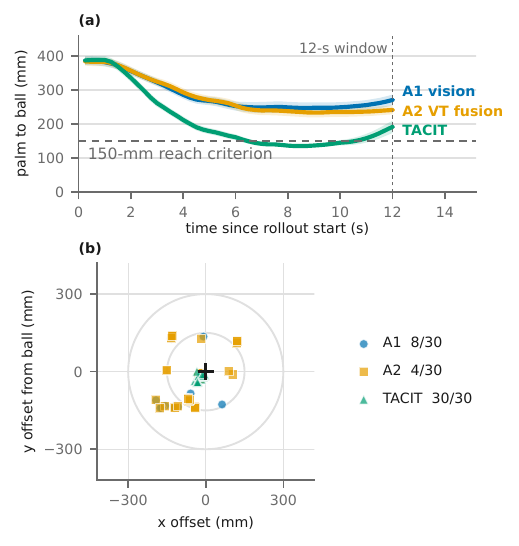}
\caption{Ball approach over thirty trials per policy. (a) Mean palm-origin-to-ball-center distance; shading is one standard error, the horizontal dashed line is the 150-mm 3D reach criterion, and the vertical dotted line marks the end of the 12-s reach-assessment window. (b) Horizontal palm offsets at closest 3D approach; rings are 150- and 300-mm references, while counts use the full 3D criterion.}
\label{fig:three_arm_approach}
\end{figure}

\textbf{The hand follows the object (Q2).} Fig.~\ref{fig:three_arm_approach} visualizes the ball trajectories and closest horizontal approach offsets across all trials; the slopes defined in Sec.~\ref{sec:evaluation-protocol} and reported in Table~\ref{tab:3} quantify this pattern on both tasks.

\begin{table}[t]
\centering
\caption{Tracking slopes $\beta$ with 95\% bootstrap intervals over placements.}
\label{tab:3}
\footnotesize
\setlength{\tabcolsep}{3pt}
\begin{tabular*}{\columnwidth}{@{}l@{\extracolsep{\fill}}ccc@{}}
\toprule
Task / axis & A1 & A2 & \textbf{TACIT} \\
\midrule
ball x & \begin{tabular}[t]{@{}c@{}}0.52\\[-1pt][-0.44, 0.70]\end{tabular} & \begin{tabular}[t]{@{}c@{}}0.02\\[-1pt][-1.82, 0.40]\end{tabular} & {\bfseries \begin{tabular}[t]{@{}c@{}}0.99\\[-1pt][0.90, 1.33]\end{tabular}} \\
ball y & \begin{tabular}[t]{@{}c@{}}0.99\\[-1pt][0.28, 2.88]\end{tabular} & \begin{tabular}[t]{@{}c@{}}0.19\\[-1pt][0.04, 0.49]\end{tabular} & {\bfseries \begin{tabular}[t]{@{}c@{}}1.08\\[-1pt][1.00, 1.13]\end{tabular}} \\
peg x & \begin{tabular}[t]{@{}c@{}}0.73\\[-1pt][-0.33, 1.01]\end{tabular} & \begin{tabular}[t]{@{}c@{}}0.16\\[-1pt][-1.64, 0.60]\end{tabular} & {\bfseries \begin{tabular}[t]{@{}c@{}}0.95\\[-1pt][0.66, 1.04]\end{tabular}} \\
peg y & \begin{tabular}[t]{@{}c@{}}0.61\\[-1pt][0.32, 0.80]\end{tabular} & \begin{tabular}[t]{@{}c@{}}-0.03\\[-1pt][-0.29, 0.35]\end{tabular} & {\bfseries \begin{tabular}[t]{@{}c@{}}0.96\\[-1pt][0.87, 1.07]\end{tabular}} \\
\bottomrule
\end{tabular*}
\end{table}

TACIT is consistent with unit slope and excludes zero on all four axes, whereas A2 excludes unit slope on all four. A1 is mixed, with broad intervals and partial tracking. Failed-trial slopes are 0.93--1.11 for TACIT and -0.10--0.10 for A2. Thus TACIT's hand position changes approximately one-for-one with object placement rather than repeating one destination. This is inconsistent with a single fixed destination, but does not exclude a coordinate-conditioned lookup over the five demonstrated regions.

\textbf{Attention before tactile activation.} Across pre-activation frames in the first 9\,s, framewise median peak error is 18, 18 and 17\,mm across the three ball seeds and 25\,mm on peg. Centroid error is larger and less stable: 17--43\,mm across ball seeds and 48\,mm on peg. These distances include placement-localization uncertainty. Thus the highest-ranked point is localized before touch, while the centroid is less consistent across seeds; its XYZ is the pooled position supplied to the policy.

\begin{figure}[t]
\centering
\includegraphics[width=\columnwidth]{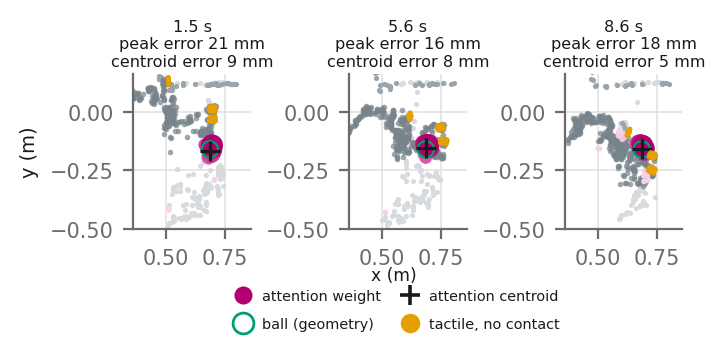}
\caption{Attention before tactile activation in one real ball rollout. The geometric ball center is independent of the head; titles show peak/centroid errors. Tactile points remain inactive until first contact at 12.6\,s.}
\label{fig:rollout_attention}
\end{figure}

In Fig.~\ref{fig:rollout_attention}, no pad activates before contact at 12.6\,s, yet the peak and centroid remain near the ball. Aggregate errors place attention near the object early, and this sequence shows it can precede tactile activation.

\subsection{Simulation Results and Held-Out Diagnostics}\label{sec:simulation-results}

\textbf{Cross-task performance (Q3).} Table~\ref{tab:4} reports task-specific simulation policies with a different embodiment and synthetic contact signal.

\begin{table}[t]
\centering
\caption{Simulation task success across three evaluation seeds; parentheses give pooled counts.}
\label{tab:4}
\footnotesize
\setlength{\tabcolsep}{2pt}
\begin{tabular}{@{}lrrr@{}}
\toprule
Task & A1 & A2 & TACIT \\
\midrule
\begin{tabular}[t]{@{}l@{}}Fruit placement\end{tabular} & \begin{tabular}[t]{@{}c@{}}$4.9 \pm 1.3$\,\%\\[-1pt](11/225)\end{tabular} & \begin{tabular}[t]{@{}c@{}}$0.0 \pm 0.0$\,\%\\[-1pt](0/225)\end{tabular} & \begin{tabular}[t]{@{}c@{}}$4.9 \pm 2.3$\,\%\\[-1pt](11/225)\end{tabular} \\
\begin{tabular}[t]{@{}l@{}}Peg insertion\end{tabular} & \begin{tabular}[t]{@{}c@{}}$2.5 \pm 3.5$\,\%\\[-1pt](2/81)\end{tabular} & \begin{tabular}[t]{@{}c@{}}$1.2 \pm 1.7$\,\%\\[-1pt](1/81)\end{tabular} & {\bfseries\boldmath \begin{tabular}[t]{@{}c@{}}14.8\,$\pm$\,5.2\,\%\\[-1pt](12/81)\end{tabular}} \\
\begin{tabular}[t]{@{}l@{}}Cube stack\end{tabular} & \begin{tabular}[t]{@{}c@{}}$5.3 \pm 0.0$\,\%\\[-1pt](12/225)\end{tabular} & \begin{tabular}[t]{@{}c@{}}$10.7 \pm 13.2$\,\%\\[-1pt](24/225)\end{tabular} & {\bfseries\boldmath \begin{tabular}[t]{@{}c@{}}20.9\,$\pm$\,6.7\,\%\\[-1pt](47/225)\end{tabular}} \\
\begin{tabular}[t]{@{}l@{}}Bimanual handover\end{tabular} & \begin{tabular}[t]{@{}c@{}}$43.3 \pm 2.7$\,\%\\[-1pt](39/90)\end{tabular} & \begin{tabular}[t]{@{}c@{}}$50.0 \pm 5.4$\,\%\\[-1pt](45/90)\end{tabular} & \begin{tabular}[t]{@{}c@{}}$54.4 \pm 12.3$\,\%\\[-1pt](49/90)\end{tabular} \\
\bottomrule
\end{tabular}
\end{table}

TACIT improves peg and cube success, fruit ties A1 at 4.9 \% (11/225), and handover shows no clear advantage, so the effect is not uniform across tasks. Absolute completion remains low; we therefore use simulation mainly as a controlled diagnostic: 138/178 failed TACIT cube rollouts reach the second-cube stage, and 68/69 failed peg rollouts occur after the peg has moved.

\textbf{Held-out spatial interpolation (Q3).} We test whether the effect is restricted to exact demonstrated coordinates using object-interaction diagnostics on the held-out in-hull placements of Sec.~\ref{sec:evaluation-protocol} (Table~\ref{tab:5}).

\begin{table}[t]
\centering
\caption{Held-out interaction. Per policy: 144 fruit/cube and 36 peg rollouts; 16 fruit/cube and 4 peg cases (ties omitted).}
\label{tab:5}
\footnotesize
\setlength{\tabcolsep}{2pt}
\begin{tabular*}{\columnwidth}{@{}l@{\extracolsep{\fill}}rrrcc@{}}
\toprule
 & \multicolumn{3}{c}{Rollouts} & \multicolumn{2}{c}{TACIT higher/lower} \\
\cmidrule(lr){2-4}\cmidrule(lr){5-6}
Task stage & A1 & A2 & TACIT & vs. A1 & vs. A2 \\
\midrule
Fruit: orange lifted & 72 & 34 & 73 & 4 / 5 & 10 / 3 \\
Fruit: lime lifted & 42 & 26 & \textbf{85} & 13 / 3 & 14 / 1 \\
Cube: lift 1 or stack & \textbf{85} & 21 & 68 & 5 / 10 & 14 / 1 \\
Cube: stack 1 & 23 & 8 & \textbf{32} & 7 / 4 & 9 / 0 \\
Cube: lift 2 or stack & 12 & 8 & \textbf{21} & 6 / 3 & 7 / 2 \\
\midrule
Peg: moved & 20 & 15 & \textbf{25} & 1 / 0 & 2 / 0 \\
\bottomrule
\end{tabular*}
\end{table}

Pooled TACIT counts exceed A2 on all six metrics; per metric it is lower in at most 3/16 fruit/cube cases and 0/4 peg cases. Against A1, first cube is lower in 10/16 cases and other metrics in at most 5/16 (peg 0/4). For orange, TACIT is higher in four cases and lower in five despite a one-rollout pooled lead. Fruit completion is 33/144 for TACIT, 18/144 for A1, and 8/144 for A2; peg and cube completion remain low. Thus TACIT improves interaction at unseen in-hull coordinates against A2, with mixed results against DP3.

\subsection{Mechanism Controls}\label{sec:mechanism}

We compare A2 with the three attention variants of Sec.~\ref{sec:policy-setup}, using three training seeds in each domain (Table~\ref{tab:6}). A2 has no attention head; the other variants share the same attention architecture and differ in supervision.

\begin{table}[t]
\centering
\caption{Three-seed comparison on simulated peg and real ball. Counts are ordered by training seed 1/2/3.}
\label{tab:6}
\footnotesize
\setlength{\tabcolsep}{2pt}
\begin{tabular*}{\columnwidth}{@{}l@{\extracolsep{\fill}}cccc@{}}
\toprule
Measure & A2 & A2.5 & \begin{tabular}[b]{@{}c@{}}TACIT-\\contact\end{tabular} & TACIT \\
\midrule
\begin{tabular}[t]{@{}l@{}}Peg moved\\[-1pt](/81)\end{tabular} & 47, 28, 27 & 33, 35, 59 & 71, 81, 81 & 80, 77, 81 \\
\begin{tabular}[t]{@{}l@{}}Peg inserted\\[-1pt](/81)\end{tabular} & 1, 1, 2 & 3, 1, 5 & 14, 16, 17 & 12, 13, 24 \\
\begin{tabular}[t]{@{}l@{}}Ball reached\\[-1pt](/30)\end{tabular} & 4, 10, 9 & 6, 4, 3 & 29, 21, 27 & 30, 27, 26 \\
\begin{tabular}[t]{@{}l@{}}Ball placed\\[-1pt](/30)\end{tabular} & 3, 3, 7 & 2, 3, 2 & 14, 17, 24 & 20, 15, 19 \\
\bottomrule
\end{tabular*}
\end{table}

\textbf{Supervision, not capacity (Q4).} Across three training seeds on ball placement, TACIT reaches the object in 30, 27, and 26 of 30 trials, compared with 4, 10, and 9 for A2. Its task success also exceeds A2 for all three seeds on both ball and simulated peg. Mean ball success is $14.4 \pm 6.3$\,\% for A2, $7.8 \pm 1.6$\,\% for A2.5, $61.1 \pm 14.0$\,\% for TACIT-contact, and $60.0 \pm 7.2$\,\% for TACIT (population standard deviation across training seeds). A2.5's mean ball success is below A2's. Across Table~\ref{tab:6}, every A2.5 seed value is below every supervised-variant value for the same metric; most ball failures miss the approach. Thus branch capacity alone is insufficient.

\textbf{Supervision timing.} Contact-interval and pre-contact supervision show no consistent ordering across three seeds in either domain: TACIT-contact and TACIT place 55/90 and 54/90 ball trials, and neither consistently leads on peg. The controls limit total supervision-budget differences but do not fully match event-level allocation. We therefore find no evidence that pre-contact placement itself improves completion. Moving targets may require temporal compensation.

\section{Limitations}\label{sec:limitations}

\textbf{Spatial and statistical scope.} Real evaluation covers five demonstrated placement regions; held-out in-hull simulation excludes distractors and object-identity changes. The primarily coordinate-conditioned head therefore does not establish semantic object detection or generalization beyond the training workspace. Three-seed comparisons cover real ball and simulated peg, not real peg or every task. The evaluation-seed deviations in Table~\ref{tab:4} describe fixed checkpoints, not training variability.

\textbf{Task and hardware scope.} The gain concerns approach, not downstream grasping, insertion, or release, and one spatial summary per arm may not suffice later. Targets assume localized contact on rigid, compact objects; deformable or distributed contact may need another target. Without motion compensation, a supervised frame may follow an earlier contact and need not correspond to the same object. Bimanual evidence covers one task; real trials use one operator, platform, and camera placement. The evaluation-only detector uses per-placement medians under occlusion and cannot resolve trial-specific placement error. The 20-Hz recorder rejects stale streams but lacks hardware triggering and per-row skew estimates.

\section{Conclusion}\label{sec:conclusion}

TACIT converts demonstration contacts into training-only camera-point attention supervision and eliminates observed pre-arrival failures in both primary real-robot benchmarks. Across three seeds, the same branch without explicit attention supervision does not reproduce the gain. Pre-contact and contact-time labels show no consistent ordering, locating the effect in supervision rather than timing or capacity. Within the evaluated workspace, contacts improve approach grounding; held-out diagnostics show improved interaction at unseen in-hull coordinates against A2. Future work should compare tactile, kinematic, and visual localization targets and test moving targets, clutter, and downstream manipulation.

\section*{Acknowledgment}

ChatGPT, Claude, and Grammarly assisted with language editing; the authors take responsibility for all technical content.

\bibliographystyle{IEEEtran}
\bibliography{refs}

\end{document}